\documentclass[conference,a4paper]{IEEEtran}
\IEEEoverridecommandlockouts
\usepackage[hidelinks]{hyperref}
\usepackage[cmex10]{amsmath}
\usepackage{amssymb,amsfonts}
\usepackage{dblfloatfix}
\usepackage[ruled,vlined]{algorithm2e}
\usepackage{graphicx}
\graphicspath{{Figures/PDF/}{Figures/PNG/}}
\usepackage{booktabs}
\usepackage{siunitx}
\usepackage[numbers,compress]{natbib}
\usepackage{texnames}
\usepackage{bm,bbm}
\usepackage{orcidlink}

\begin{document}

\title{\uppercase{Deep Learning for Remote Sensing to Improve Flood Inundation Mapping}
}

\author{	\IEEEauthorblockN{Yogesh \ Bhattarai\orcidlink{0000-0002-0377-0823}}
	\IEEEauthorblockA{\textit{Department of Civil and Environmental Engineering}\\
		Howard University\\
        Washington, D.C, 20060, USA.\\
		yogesh.bhattarai@bison.howard.edu}
	\and
	\IEEEauthorblockN{Vijay \ Chaudhary}
	\IEEEauthorblockA{\textit{Department of Computer Science and Electrical Engineering}\\
		Howard University\\
        Washington, D.C, 20060, USA.\\
		vijay.chaudhary@howard.edu}
    \and
	\IEEEauthorblockN{Wai Lim \ Ku}
	\IEEEauthorblockA{\textit{Department of Medicine}\\
		Howard University\\
        Washington, D.C, 20060, USA.\\
		Wailim.Ku@howard.edu}
	\and
	\IEEEauthorblockN{Sanjib \ Sharma\orcidlink{0000-0003-2735-1241}}
	\IEEEauthorblockA{\textit{Department of Civil and Environmental Engineering}\\
		Howard University\\
        Washington, D.C, 20060, USA.\\
		sanjib.sharma@howard.edu}
}

\maketitle
\begin{abstract}
	Flooding is the most pervasive natural disaster worldwide. Timely and accurate flood inundation mapping are essential for informing disaster risk management. Optical satellite missions provide high-resolution, multispectral observations critical for flood detection and inundation mapping. However, their operational utility is severely constrained by cloud cover during extreme precipitation events. Conventional cloud-removal techniques based on temporal compositing or interpolation often fail to capture inundation dynamics. In this study, we introduce a cloud-removal framework for flood imagery based on Denoising Diffusion Probabilistic Models, leveraging the Masked Diffusion Transformer architecture. The proposed approach exploits self-attention mechanisms to capture wider spatial context and employs masked token modeling to explicitly learn the reconstruction of cloud-obscured regions. Trained on multispectral Sentinel-2B flood scenes with realistic cloud patterns, the model generates cloud-free image realizations that preserve both visual fidelity and hydrological consistency. Reconstruction performance is evaluated using standard image quality metrics alongside flood-specific hydrological measures, demonstrating improved continuity of water bodies and preservation of spectral signatures critical for water detection indices. The results indicate that diffusion-based generative modeling offers a robust and physically consistent alternative for cloud removal in optical flood monitoring, enabling more reliable, continuous observations to support disaster risk management and flood-related decision making.

\end{abstract}

\begin{IEEEkeywords}
	Flooding, satellite imagery, diffusion models, machine learning
\end{IEEEkeywords}

\section{Introduction}

\noindent Flooding is the most pervasive and widely disruptive natural hazard, affecting billions of people worldwide and causing over \$100 billion in annual economic losses\cite{he_mobility_2022}. Remote sensing has emerged as the critical tool for large-scale flood monitoring, revealing inundated areas that is difficult to achieve through ground-based surveys\cite{pekel2016high, portales2023global}. The European Space Agency's Copernicus Sentinel-2 mission provides valuable multispectral information on flood detection and mapping\cite{tarpanelli2022effectiveness,farhadi2025introducing}. In particular, Sentinel-2B offers high temporal resolution (5-day revisit time at the equator when combined with Sentinel-2A) and spatial resolutions ranging from 10m to 60m\cite{spoto2012overview}. These characteristics make it well-suited for resolving detailed hydrological features and delineating flood extents using spectral indices\cite{drusch2012sentinel}.\\
\noindent Sentinel-2B optical imagery faces inherent limitations during extreme precipitation events because passive sensors cannot penetrate cloud cover. In tropical and subtropical regions, more than 70\% of optical satellite imagery becomes unusable during flooding seasons\cite{whitcraft2015cloud}. This results in critical information gaps when timely flood extent estimates are most needed for emergency response. Synthetic Aperture Radar (SAR) sensors such as Sentinel-1C can penetrate clouds and operate in all weather conditions\cite{torres2017sentinel}; however, speckle noise\cite{sebastianelli2022speckle}  and reduced interpretability in complex urban environments limit their reliability for detailed inundation mapping\cite{koppel2017sensitivity}. 
Robust reconstruction of cloud-obscured regions in optical flood imagery remains essential for maintaining continuous, high-quality flood observation records.\\
\noindent Classical approaches to cloud removal relies primarily on temporal compositing\cite{fraser2009method}, mosaicking\cite{kang2016automatic}, or multi-temporal interpolation\cite{hu2020multi}, where cloudy pixels are replaced with data from cloud-free acquisitions on different dates. However, these methods are fundamentally limited in flood monitoring applications because floodwaters are highly dynamic\cite{shen2015missing}. A clear image acquired days before or after a flood event does not accurately reflect the current inundation state, as water levels can change dramatically within hours. Temporal interpolation methods may fail to capture the peak extent of flooding or misrepresent the spatial distribution of inundated areas. 
More recently, advances in deep learning have significantly enhanced cloud removal techniques, particularly through the use of Generative Adversarial Networks (GANs) for image inpainting and reconstruction of cloud-obscured regions\cite{edirisinghe2025generative, darbaghshahi2021cloud,zhao2021cloud}. However, GANs are often prone to training instability and mode collapse, where the generator learns to produce only a limited subset of possible outputs\cite{barsha2025depth}. In flood mapping applications, such failures can be particularly misleading, resulting in the erroneous reconstruction of dry land in inundated areas or the artificial disconnection of river channels. Training instability and mode collapse often leads to failure of GANs to preserve hydrological continuity and physical consistency required for quantitative analysis and emergency management applications\cite{singh2018cloud}.\\
\noindent In this study, we introduce a generative approach based on Denoising Diffusion Probabilistic Models\cite{ho2020denoising}, specifically leveraging the Masked Diffusion Transformer (MDT) architecture\cite{gao2023masked}. Diffusion models have recently surpassed GANs in image generation quality through iterative reversal of a gradual noising process \cite{ho2020denoising}, offering improved stability and more complete mode coverage. We hypothesize that MDT is particularly well suited for cloud removal in flood imagery for three reasons. First, unlike conventional CNN-based U-Net architectures, the Transformer backbone utilizes self-attention mechanisms to capture global spatial context across the entire image. This enables the model to infer the presence and characteristics of floodwater beneath cloud cover by analyzing the trajectory, extent, and spectral properties of visible water in surrounding unclouded regions.Second, MDT employs masked token modeling during training, explicitly learning to predict missing image patches\cite{gao2023masked}. This paradigm aligns naturally with the physical problem of cloud obstruction, where clouds act as occlusions that mask the underlying ground surface. Third, diffusion models generate predictions through iterative refinement, progressively improving reconstruction quality and maintaining better consistency with physical constraints compared to single-step generators.\\ 
\noindent The objective of this study is to develop and evaluate an MDT-based approach for generating cloud-free multispectral realizations of Sentinel-2B flood imagery that preserves visual clarity and hydrological accuracy. The key contributions include: (i) adapting the MDT architecture for multispectral satellite imagery reconstruction with explicit consideration of flood-specific spectral characteristics; (ii) training and validating the model using Sentinel-2B flood scenes with  realistic cloud cover; (iii) evaluating reconstruction performance using standard image quality and hydrological consistency metrics; and (iv) demonstrating the model's ability to maintain spatial continuity of water bodies and preserve spectral signatures critical for water detection indices. The insights gained from this study can produce more reliable flood monitoring under cloudy conditions that is crucial for disaster risk management.
\section{Methods and Materials}
\subsection{Study Data}
\noindent We utilize post-2017 Sentinel-2B multispectral imagery at 10m spatial resolution covering nine flood events across nine different locations for model development. For each event, a 50-day temporal window centered on the flood occurrence is considered, spanning incidents from 2018 to 2025. Due to temporal variation of Sentinel-2B, we could only manage total of 62 images for training and validation of model. Each image is paired with a corresponding classification map delineating five land cover classes: Invalid (class 1), Land (class 2), Water (class 3), Cloud (class 4), and Flooded areas (class 5), along with a Digital Elevation Model (DEM). Pixel values are normalized to the range [0, 1] using per-channel statistics computed across the training dataset and extract non-overlapping patches of 256 × 256 pixels from both imagery and classification maps to facilitate efficient training while maintaining spatial context. We apply geometric transformations (horizontal/vertical flipping and 90° rotations, each with 50\% probability) to both satellite images and classification maps to preserve spatial correspondence.

\subsection{Model Architecture}

\noindent We develop a MDT modeling scheme for joint cloud detection and inpainting in satellite imagery. Following the MDT framework of Gao et al.\cite{gao2023masked}, we adapt the architecture specifically for multispectral satellite applications. We integrate diffusion-based generative model with transformer attention mechanisms. The model comprises three primary components: an encoder network processing all input tokens, an asymmetric decoder network processing only masked tokens, and a separate cloud detection head for semantic segmentation.

\noindent The model accepts multispectral satellite images with dimensions $(B, 10, 256, 256)$ and DEM features with dimensions $(B, 3, 256, 256)$. We divide input images into non-overlapping $16 \times 16$ pixel patches, yielding 256 patches in a $16 \times 16$ grid. These patches are projected into 768-dimensional token embeddings through $\text{Conv2d}(10, 768, k=16, s=16)$, and augmented with learnable positional embeddings encoding spatial relationships.

\noindent The DEM encoder provides critical terrain context for generating water bodies and flooded regions. We process elevation data through three parallel channels. These channels are stacked and projected through a convolutional neural network to produce DEM embeddings aligned dimensionally with patch embeddings. The encoder processes all tokens through 12 transformer blocks with 768-dimensional hidden size, 12 attention heads, MLP expansion ratio of 4.0, and dropout probability of 0.1. Each block employs Adaptive Layer Normalization (AdaLN-Zero) conditioning to incorporate diffusion timestep information through six modulation parameters.

\noindent The decoder processes only masked tokens while attending to unmasked tokens as context which reduces computational cost. We employ 4 decoder blocks maintaining 768 dimensions and 12 attention heads. Each block performs self-attention on masked tokens, followed by cross-attention where masked tokens query unmasked context via $\text{softmax}(QK^T / \sqrt{d})V$ with $Q$ from masked tokens and $K, V$ from unmasked tokens. Masked patches use a learnable token initialized from a truncated normal distribution with standard deviation 0.02, replacing cloud-affected patches as $\text{masked\_tokens} = \text{tokens} \cdot (1 - \text{mask}) + \text{mask\_token} \cdot \text{mask}$. The cloud detection head comprises 4 transformer encoder layers processing the full encoded representation, with output projected through a linear layer mapping 768-dimensional features to 1280 values, reshaped into segmentation logits of dimension $(B, 5, 256, 256)$ covering invalid pixels, land, water, cloud, and flooded areas. 

\subsection{Loss Function}
We employ a combined loss addressing both segmentation and reconstruction tasks:
\begin{equation}
\mathcal{L}_{\text{total}} = \mathcal{L}_{\text{segmentation}} + \mathcal{L}_{\text{reconstruction}}
\end{equation}
Our segmentation map has five landcover classes. We observe imbalanced distributions for different classes as land (class 2) dominates over other classes. We combine three complementary losses addressing multi-class segmentation with imbalanced distributions:
\begin{equation}
\mathcal{L}_{\text{seg}} = \lambda_{CE} \mathcal{L}_{CE} + \lambda_{Dice} \mathcal{L}_{Dice} + \lambda_{Focal} \mathcal{L}_{Focal}
\end{equation}
\noindent where $\lambda_{CE} = 1.0$ handles standard classification, $\lambda_{Dice} = 0.5$ addresses class imbalance, and $\lambda_{Focal} = 0.5$ focuses on challenging examples. Cross-entropy loss measures pixel-level accuracy:
\begin{equation}
\mathcal{L}_{CE} = -\frac{1}{N} \sum_{i=1}^{N} \sum_{c=1}^{C} y_{i,c} \log(p_{i,c})
\end{equation}
\noindent where $y_{i,c}$ is the one-hot target and $p_{i,c}$ is predicted probability. We select dice loss for optimizing spatial overlap:
\begin{equation}
\mathcal{L}_{Dice} = 1 - \frac{2 |P \cap G| + \epsilon}{|P| + |G| + \epsilon}
\end{equation}
\noindent where $P$ and $G$ represent prediction and ground truth sets, with $\epsilon = 10^{-6}$ preventing division by zero. Focal loss down-weights well-classified examples:
\begin{equation}
\mathcal{L}_{Focal} = -\frac{1}{N} \sum_{i=1}^{N} \alpha_i (1 - p_i)^{\gamma} \log(p_i)
\end{equation}
\noindent where $\gamma = 2.0$ controls focusing strength. For reconstruction, we combine three complementary losses to reduce training fluctuations and improve inpainting quality:
\begin{equation}
\mathcal{L}_{\text{recon}} = \lambda_{MSE} \mathcal{L}_{MSE} + \lambda_{Perceptual} \mathcal{L}_{Perceptual} + \lambda_{SSIM} \mathcal{L}_{SSIM}
\end{equation}
\noindent where $\lambda_{MSE} = 0.6$ ensures pixel-wise accuracy, $\lambda_{Perceptual} = 0.4$ captures texture similarity, and $\lambda_{SSIM} = 0.1$ preserves structural consistency. Mean squared error provides stable gradients on masked regions:
\begin{equation}
\mathcal{L}_{MSE} = \frac{1}{|M|} \sum_{i \in M} (\hat{x}_i - x_i)^2
\end{equation}
\noindent where $M$ denotes masked pixels, $\hat{x}_i$ is the prediction, and $x_i$ is the original value. Perceptual loss compares multi-scale feature representations:
\begin{equation}
\mathcal{L}_{Perceptual} = \sum_{l} \frac{1}{|M_l|} \| \phi_l(\hat{x}) \odot M_l - \phi_l(x) \odot M_l \|_2^2
\end{equation}
\noindent where $\phi_l$ extracts features at layer $l$ of an encoder, and $M_l$ is the mask downsampled to the corresponding resolution. Structural similarity index loss maintains local patterns:
\begin{equation}
\mathcal{L}_{SSIM} = 1 - \frac{(2\mu_{\hat{x}}\mu_x + C_1)(2\sigma_{\hat{x}x} + C_2)}{(\mu_{\hat{x}}^2 + \mu_x^2 + C_1)(\sigma_{\hat{x}}^2 + \sigma_x^2 + C_2)}
\end{equation}
\noindent where $\mu$ and $\sigma^2$ denote local means and variances computed with an $11 \times 11$ Gaussian window, $\sigma_{\hat{x}x}$ is the local covariance, and $C_1 = 0.0001$, $C_2 = 0.0009$ are stability constants. This combination addresses limitations of single-metric losses: MSE provides smooth optimization gradients, perceptual loss prevents blurry outputs by enforcing feature-level similarity, and SSIM preserves edges and structural coherence in the reconstructed regions.
\subsection{Training Procedure}
We train the model using AdamW optimizer with initial learning rate $1 \times 10^{-4}$, weight decay 0.01, and batch size 4. Learning rate scheduling via ReduceLROnPlateau reduces the rate by 0.5 with patience of 20 epochs when validation loss plateaus. We apply gradient clipping with maximum norm 1.0 and train for 100 epochs maximum with early stopping after 20 epochs without improvement, retaining the checkpoint achieving minimum validation loss.

The diffusion process implements forward diffusion parameterized as:
\begin{equation}
q(x_t|x_0) = \mathcal{N}(x_t; \sqrt{\bar{\alpha}_t}x_0, (1-\bar{\alpha}_t)I)
\end{equation}
 where $\bar{\alpha}_t = \prod_{i=1}^{t} \alpha_i$ and $\alpha_t = 1 - \beta_t$. We employ 1000 timesteps with linear noise schedule from $\beta_{\text{start}} = 0.0001$ to $\beta_{\text{end}} = 0.02$. During training, we randomly sample timestep $t$ and add Gaussian noise as $x_t = \sqrt{\bar{\alpha}_t} x_0 + \sqrt{1 - \bar{\alpha}_t} \epsilon$ where $\epsilon \sim \mathcal{N}(0, I)$. During inference, reverse diffusion iteratively denoises masked tokens through 50 steps, maintaining unmasked tokens fixed as $x_t = x_{\text{original}} \cdot (1 - \text{mask}) + x_t^{\text{denoised}} \cdot \text{mask}$ to ensure only cloud-affected regions undergo regeneration while visible areas remain unchanged.

\subsection{Evaluation Metrics}

We assess performance using complementary metrics capturing segmentation quality and reconstruction fidelity. Intersection over Union for each class $c$ is:

\begin{equation}
\text{IoU}_c = \frac{|P_c \cap G_c|}{|P_c \cup G_c|}
\end{equation}

\noindent where $P_c$ and $G_c$ denote predicted and ground truth pixels. Mean IoU serves as our primary metric:

\begin{equation}
\text{mIoU} = \frac{1}{C} \sum_{c=1}^{C} \text{IoU}_c
\end{equation}

\section{Results}
\noindent We monitor the training dynamics of MDT model for cloudy pixel segmentation and reconstruction for over 100 epochs (Figure 1). Figure 1a exhibits a consistent decrease in loss for both the training and validation sets. This indicates stable convergence without significant overfitting for the model. The close alignment between the training and validation curves indicate strong generalization on Sentinel-2B spatiotemporal data. Figure 1b illustrates rapid learning during the segmentation process, with the validation loss decreasing from 1.2 to 0.55. This indicates the encoder effectively captures high-level semantic features required for delineating flood boundaries. Figure 1c represents the reconstruction loss from masked diffusion process, where the validation loss stabilizes from 0.2 to 0.05. This trend demonstrates the model’s ability to restore masked patches while effectively incorporating DEM elevation data maintaining reliable classification performance up to approximately 60-70\% cloudy pixel coverage per scene. Figure 1d highlights the generalization capability of the MDT model, with the Intersection over Union (IoU) increasing from 0.3 to a peak value of 0.57 around epoch 80. The improvement in IoU suggests that the attention mechanisms are effective in resolving complex flood patterns and distinguishing water bodies from cloud shadows.
\begin{figure}
    \centering
    \includegraphics[width=0.8\linewidth]{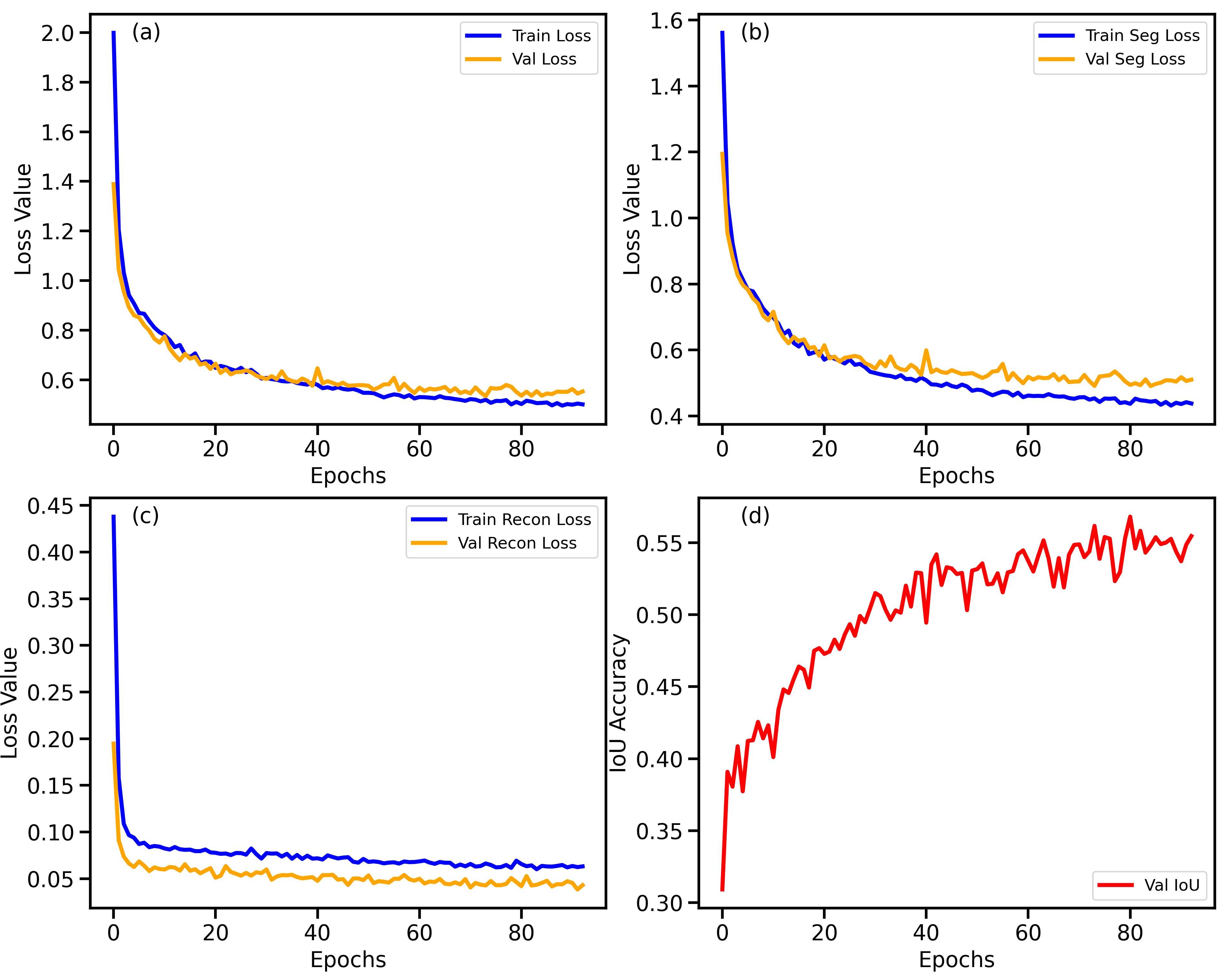}
    \caption{Training and validation performance metrics for the Masked Diffusion Transformer (MDT) for flood mapping: (a) total loss showing overall convergence of the segmentation and reconstruction tasks; (b) segmentation loss illustrating the model's ability to classify flood pixels; (c) reconstruction loss tracks the fidelity of the generative component in restoring masked/cloudy patches; and (d) validation IoU demonstrates the progressive improvement in spatial accuracy.}
    \label{metrics}
\end{figure}
\\
We apply the trained MDT model to predict flooded areas during the 2021 Tennessee flood event \cite{gangrade2023unraveling}. The 2021 Central Tennessee flood was a catastrophic flash flood triggered by extreme precipitation on August 21, 2021, which severely impacted the City of Waverly and surrounding areas. The event resulted in 21 fatalities and caused significant damage to hundreds of homes and properties\cite{gangrade2023unraveling}. Figure 2a represents the raw satellite image from the flooding events. Figure 2b represents the detected cloudy pixel. We observe the cloudy pixel covers 7.91\% of the selected image. Figure 2c represents the reconstructed water bodies and mapped flooded areas. We observe model detects 0.53\% as the flooded area and reconstructs cloudy pixel distribution from 7.91\% to 6.09\%. This is major improvement over the traditional threshold-based cloud masking methods like Sen2Cor for Sentinel-2 which typically retain 10-15\% residual cloud contamination under complex cloud conditions. Simpler CNN-based classifiers such as U-Net trained on the same dataset achieved approximately 8.5\% residual cloud coverage.  

\begin{figure}[ht]
    \centering
    \includegraphics[width=0.77\linewidth]{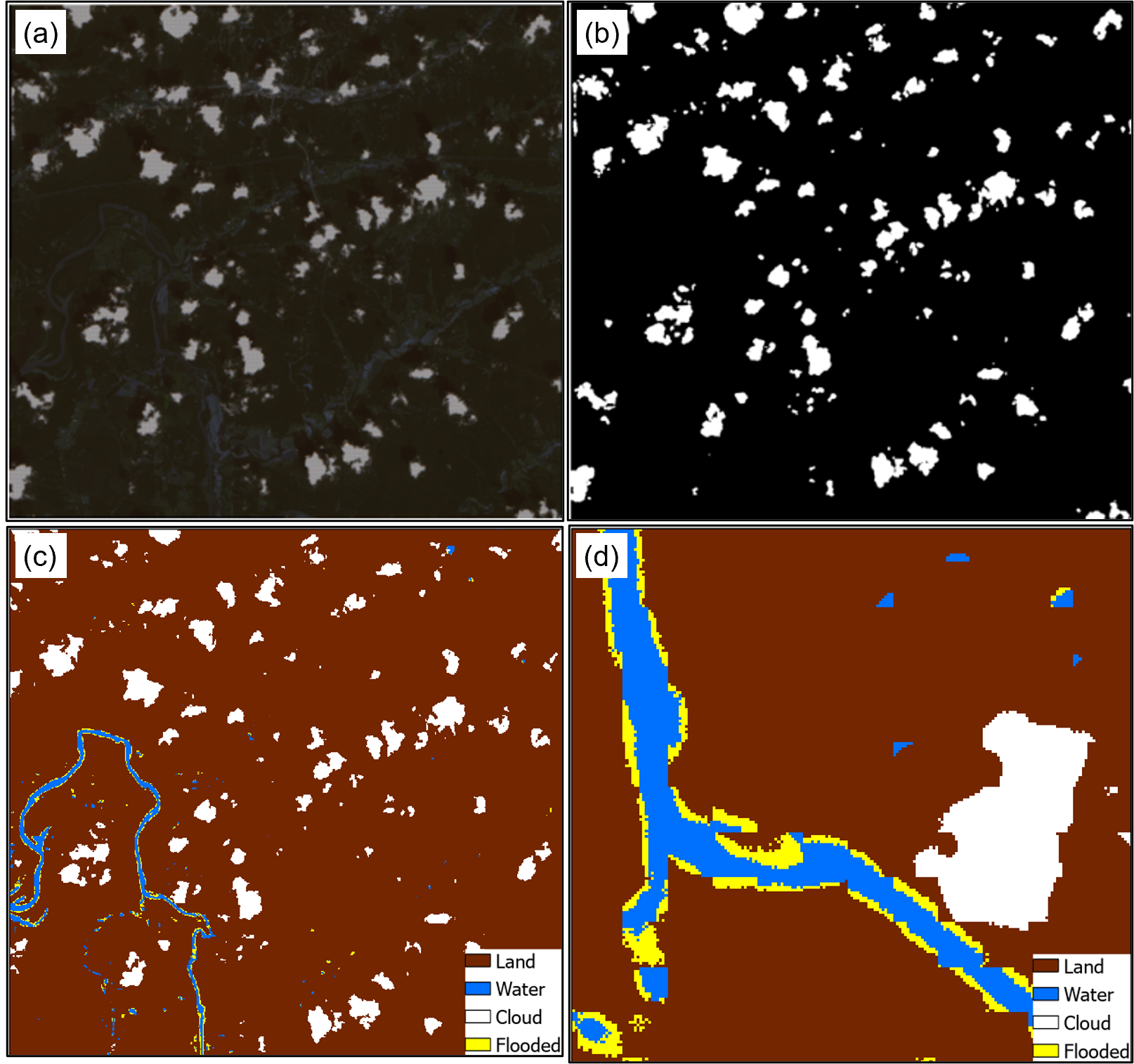}
    \caption{Application of the proposed MDT model on the August 2021 Tennessee flood event: (a) raw Sentinel-2B satellite imagery; (b) cloudy pixels identified by the model; (c) final output showing reconstructed water bodies and mapped flood inundation extents; and (d) zoomed section depicting reconstructed water bodies and flood extents.}
    \label{fig:placeholder}
\end{figure}
\section{Discussion and Conclusion}
\noindent Our study demonstrates that MDT enhances flood prediction capabilities during extreme precipitation events characterized by extensive cloud coverage. By integrating diffusion models with earth observation, MDT mitigates cloud obstruction during flooding events. MDT generates physically plausible reconstructions using spatial context and learned patterns rather than relying on temporal compositing or multispectral analysis. This addresses a critical operational challenge in disaster response, as cloud cover creates temporal gaps in earth observation when flood information is most urgently needed. While satellite flood mapping typically requires clear conditions or SAR sensors, our approach enables the use of high resolution optical imagery under adverse atmospheric conditions, potentially accelerating damage assessment and emergency response timelines.\\
\noindent We acknowledge several limitations. The model was trained using a limited number of localized flood events, which may affect generalizability across regions with different topographic, climatic, or land cover conditions. Limited ground-truth validation data also introduce uncertainty. Elevation data are incorporated to help reduce hallucination artifacts; however, the model relies primarily on remotely sensed imagery and does not explicitly incorporate physics-based processes. Also, our model is proposed as cloud-aware optical method focusing on cloud presence only. Future work may benefit from expanding the training dataset to include diverse flood scenarios. Incorporating additional physical constraints beyond elevation, such as drainage network topology and antecedent soil moisture conditions, could further support physical consistency. Integrating physics-based hydrodynamic model with fused optical and SAR data, and explicitly modeling temporal flood dynamics may further improve predictive robustness. Future research can benefit from adding the different cloud types like thin uniform cloud versus broken cloud structure with varying spectral contrast. Additionally, comparison of model performance with against Sen2Cor (rule-based)\cite{aybar2022cloudsen12}, a standard U-Net\cite{weng2019unet}, and a Vision Transformer (ViT)\cite{wang2022vit} baseline trained under identical conditions will improve the trustworthiness. 

\bibliographystyle{IEEEtranN}
\bibliography{references}

\end{document}